\documentclass{article}

\usepackage{microtype}
\usepackage{amsmath}
\usepackage{amsfonts}
\usepackage{graphicx}
\usepackage{subfigure}
\usepackage{booktabs} % for professional tables
\usepackage{csquotes}
\usepackage{siunitx}
\usepackage{dcolumn}

\usepackage{hyperref}

\usepackage[accepted]{icml2018}
\gdef\isarxiv{1}

\usepackage{ifthen}
\usepackage{xcolor}

\newcommand*{\todo}[2][]{\textcolor{red}{[\textbf{\ifthenelse{\equal{#1}{}}{TODO}{TODO(#1)}}: #2]}}

\newcommand{\attack}{BPDA}

\newcommand{\attacklong}{Backward Pass Differentiable Approximation}

\newcommand{\github}{
    \tiny{\url{https://github.com/anishathalye/obfuscated-gradients}}
}

\newcommand*{\supplement}{\ifdefined\isarxiv{Appendix}\else{Supplement Section}\fi}

\newcommand{\cifar}{\mbox{CIFAR-10}}

\icmltitlerunning{Obfuscated Gradients Give a False Sense of Security:
Circumventing Defenses to Adversarial Examples}

\begin{document}

\twocolumn[
  \icmltitle{Obfuscated Gradients Give a False Sense of Security:\\
  Circumventing Defenses to Adversarial Examples}



\icmlsetsymbol{equal}{*}

\begin{icmlauthorlist}
\icmlauthor{Anish Athalye}{equal,mit}
\icmlauthor{Nicholas Carlini}{equal,ucb}
\icmlauthor{David Wagner}{ucb}
\end{icmlauthorlist}

\icmlaffiliation{mit}{Massachusetts Institute of Technology}
\icmlaffiliation{ucb}{University of California, Berkeley}

\icmlcorrespondingauthor{Anish Athalye}{aathalye@mit.edu}
\icmlcorrespondingauthor{Nicholas Carlini}{npc@berkeley.edu}

\icmlkeywords{Machine Learning, ICML}

\vskip 0.3in
]



\printAffiliationsAndNotice{\icmlEqualContribution} % otherwise use the standard text.

\begin{abstract}
    We identify
obfuscated gradients, a kind of gradient masking,
as a phenomenon that leads to a false sense of
security in defenses against adversarial examples.
While defenses that cause obfuscated gradients appear to defeat
iterative optimization-based attacks,
we find defenses relying on this effect can be
circumvented.
We describe characteristic behaviors of defenses exhibiting the effect,
and for each of the three types of obfuscated gradients we discover,
we develop attack techniques to overcome it.
In a case study, examining non-certified white-box-secure defenses at ICLR 2018,
we find obfuscated gradients are a common
occurrence, with 7 of 9 defenses relying on obfuscated
gradients. Our new attacks successfully circumvent
6 completely, and 1 partially, in the original threat model each paper considers.

% right now, the only thing our abstract doesn't mention is our discussion
% section (the "security 101") but I think that's fine; that's not really a
% novel contribution, it's just something that's being reiterated for the nth
% time

\end{abstract}

\vspace{-2em}
\section{Introduction}
\label{sec:introduction}

In response to the susceptibility of neural networks to adversarial examples
\cite{szegedy2013intriguing,biggio2013evasion},
there has been significant interest recently in
constructing defenses to increase the robustness of neural networks. While
progress has been made in understanding and defending against adversarial
examples in the white-box setting, where the adversary has full access to
the network, a complete solution has not yet been found. %To the best of our
%knowledge, all defenses against adversarial examples in the white-box setting
%published in peer-reviewed
%venues to
%date~\cite{papernot2016distillation,hendrik2017detecting,hendrycks2017early,meng2017magnet,zantedeschi2017efficient}
%are vulnerable to powerful iterative optimization-based
%attacks~\cite{sp2017:carlini,carlini2017magnet,carlini2017adversarial}.

As benchmarking against iterative optimization-based attacks (e.g.,
\citet{kurakin2016adversarial,madry2018towards,sp2017:carlini})
has become standard practice in evaluating
defenses, new defenses have arisen that appear to be robust against these
powerful optimization-based attacks. %These defenses sometimes exhibit
%surprising behavior such as black-box attacks or single-step attacks like FGSM
%achieving better performance than iterative white-box attacks, but in general,
%they seem to provide a much greater degree of robustness than previous
%approaches.

We identify one common reason why many defenses provide apparent robustness
against iterative optimization attacks: \emph{obfuscated gradients}, a term we
define as a special case of gradient masking \cite{papernot2017blackbox}.
Without a good gradient, where following the gradient does not successfully
optimize the loss, iterative optimization-based methods cannot succeed.
We identify three types of obfuscated gradients:
\emph{shattered gradients} are nonexistent or incorrect gradients
caused either intentionally through non-differentiable operations or
unintentionally through numerical instability;
\emph{stochastic gradients} depend on test-time randomness;
and \emph{vanishing/exploding gradients} in very deep
computation result in an unusable gradient.

We propose new techniques to overcome obfuscated gradients caused by these
three phenomena. We address gradient shattering
with a new attack technique we call \attacklong, where we
approximate derivatives by computing the forward pass
normally and computing the backward pass using a
differentiable approximation of the function. We compute gradients of
randomized defenses by applying Expectation Over
Transformation~\cite{athalye2017synthesizing}. We solve vanishing/exploding
gradients through reparameterization and optimize over a space where
gradients do not explode/vanish.

To investigate the prevalence of obfuscated gradients and understand the
applicability of these attack techniques,
we use as a case study the ICLR 2018 non-certified
defenses that claim white-box robustness.
We find that obfuscated gradients are a common occurrence, with
7 of 9 defenses relying on this phenomenon. Applying the new
attack techniques we develop, we overcome obfuscated gradients and
circumvent 6 of them completely, and 1 partially, under the original
threat model of each paper. Along with this, we offer an analysis of the
evaluations performed in the papers.

Additionally,
we hope to provide researchers with a common baseline of knowledge,
description of attack techniques, and common evaluation pitfalls, so that
future defenses can avoid falling vulnerable to these same attack approaches.

To promote reproducible research, we release our re-implementation of each of
these defenses, along with implementations of our attacks for
each.~\footnote{\github{}}

\section{Preliminaries}
\label{sec:prelim}

\subsection{Notation}
We consider a neural network $f(\cdot)$ used for classification where
$f(x)_i$ represents the probability that image $x$ corresponds to label $i$.
We classify images, represented as $x \in [0,1]^{w \cdot h \cdot c}$ for a
$c$-channel image of width $w$ and height $h$.
We use $f^j(\cdot)$ to refer to layer $j$ of the neural network,
and $f^{1..j}(\cdot)$ the composition of layers $1$ through $j$.
We denote the classification of the network as
$c(x)=\text{arg max}_i f(x)_i$, and
$c^*(x)$ denotes the true label.

\subsection{Adversarial Examples}
Given an image $x$ and classifier $f(\cdot)$, an adversarial example
\cite{szegedy2013intriguing} $x'$ satisfies two properties: $\mathcal{D}(x,x')$
is small for some distance metric $\mathcal{D}$, and $c(x') \ne c^*(x)$.
That is, for images, $x$ and $x'$ appear visually similar but $x'$ is
classified incorrectly.

In this paper, we use the $\ell_\infty$ and $\ell_2$ distortion metrics
to measure similarity. Two images which have a small distortion
under either of these metrics will appear visually identical.
We report $\ell_\infty$ distance in the normalized $[0,1]$ space,
so that a distortion of $0.031$ corresponds to $8/256$,
and $\ell_2$ distance as the total root-mean-square distortion
normalized by the total number of pixels (as is done in prior work).
%\[ \ell_2(x,x') = {1 \over N} \cdot \left(\sum\limits_{i=0}^N (x_i-x'_i)^2\right)^{1\over2} \]
%FIXME can cut above for space
% XXX above, 1/N should be inside square root

\subsection{Datasets \& Models}
\label{sec:datasets}
We evaluate these defenses on the same datasets
on which they claim robustness.

% MNIST \cite{lecun1998mnist} for
%\citet{samangouei2018defensegan}, CIFAR-10 \cite{krizhevsky2009learning} for
%\citet{madry2018towards,song2018pixeldefend,ma2018characterizing,buckman2018thermometer,dhillon2018stochastic},
%and ImageNet \cite{krizhevsky2012imagenet} for
%\citet{guo2018countering,xie2018mitigating}.
If a defense argues security on MNIST and any other dataset, we only evaluate
the defense on the larger dataset. On MNIST and \cifar{}, we evaluate defenses
over the entire test set and generate untargeted adversarial examples. On
ImageNet, we evaluate over 1000 randomly selected images in the test set,
construct \emph{targeted} adversarial examples with randomly selected target
classes, and report attack success rate in addition to model accuracy.
Generating targeted adversarial examples is a strictly harder problem that we
believe is a more meaningful metric for evaluating attacks.~\footnote{Misclassification is a less meaningful
  metric on ImageNet, where a misclassification of closely related classes (e.g., a German shepherd classified as a Doberman) may not be meaningful.}
Conversely, for a defender, the harder task is to argue robustness to
untargeted attacks.

We use standard models for each dataset.
For MNIST we use a standard 5-layer convolutional neural network which reaches
$99.3\%$ accuracy. On \cifar{} we train a wide ResNet
\cite{zagoruyko2016wide,he2016deep} to $95\%$ accuracy.
For ImageNet we use the InceptionV3
\cite{szegedy2016rethinking} network which
reaches $78.0\%$ top-1 and $93.9\%$ top-5 accuracy.

\subsection{Threat Models}

Prior work considers adversarial examples in white-box and black-box threat models.
In this paper, we consider defenses designed for the \emph{white-box} setting, where
the adversary has full access to the neural network classifier (architecture
and weights) and defense, but not test-time randomness (only the distribution).
We evaluate each defense under the threat model under which it claims to be secure
(e.g., bounded $\ell_\infty$ distortion of $\epsilon = 0.031$).
It often easy to find imperceptibly perturbed adversarial examples by violating
the threat model, but by doing so under the original threat model, we show that
the original evaluations were inadequate and the claims of defenses' security
were incorrect.

\subsection{Attack Methods}
We construct adversarial examples with iterative optimization-based
methods. For a given instance $x$,
these attacks attempt to search for a
$\delta$ such that $c(x+\delta) \ne c^*(x)$ either minimizing $\|\delta\|$,
or maximizing the classification loss on $f(x+\delta)$.
To generate $\ell_\infty$ bounded adversarial examples we use Projected
Gradient Descent (PGD) confined to a specified $\ell_\infty$ ball; for
$\ell_2$, we use the Lagrangian relaxation of \citet{sp2017:carlini}.
We use between 100 and 10,000 iterations of gradient descent, as needed to
obtain convergance.
The specific choice of optimizer is far less important than choosing to use
iterative optimization-based methods~\cite{madry2018towards}.

\section{Obfuscated Gradients}
\label{sec:obfuscated}

A defense is said to cause \emph{gradient masking} if it
``does not have useful gradients'' for generating adversarial
examples~\cite{papernot2017blackbox};
gradient masking is known to be an incomplete defense to adversarial
examples~\cite{papernot2017blackbox,tramer2018ensemble}.
Despite this, we observe that 7 of the ICLR 2018 defenses rely on this
effect.

To contrast from previous defenses which cause gradient masking
by learning to break gradient descent (e.g., by learning to make
the gradients point the wrong direction \cite{tramer2018ensemble}),
we refer to the case
where defenses are \emph{designed} in such a way that the
constructed defense necessarily causes gradient masking
as \emph{obfuscated gradients}.
We discover three ways in which defenses obfuscate gradients (we use
this word because in these cases, it is
the defense creator who has obfuscated the gradient information);
we briefly define and discuss each of them.

\textbf{Shattered Gradients} are caused when a defense is non-differentiable,
introduces numeric instability, or otherwise causes a gradient
to be \emph{nonexistent or incorrect}.
Defenses that cause gradient shattering can do so unintentionally, by
using differentiable operations but where following the gradient
does not maximize classification loss globally.

\textbf{Stochastic Gradients} are caused by randomized defenses, where
either the network itself is randomized or the input is randomly transformed before
being fed to the classifier, causing the gradients to become randomized.
This causes methods using a single
sample of the randomness to incorrectly estimate the true gradient.% direction
%and fail to converge to a point that maximizes the expected classification loss
%of the randomized classifier.

\textbf{Exploding \& Vanishing Gradients} are often caused by defenses that
consist of multiple iterations of neural network evaluation, feeding the
output of one computation as the input of the next. %For example,
%Defense-GAN~\cite{samangouei2018defensegan} projects an input point onto the
%manifold of a GAN via gradient descent.
%
This type of computation, when unrolled, can be viewed as an extremely deep
neural network evaluation, which can cause vanishing/exploding
gradients.%~\cite{bengio1994longterm}.

\subsection{Identifying Obfuscated \& Masked Gradients}

Some defenses intentionally break gradient
descent and cause obfuscated gradients.
However, others defenses \emph{unintentionally} break gradient descent,
but the
cause of gradient descent being broken is a direct result of
the design of the neural network.
We discuss below characteristic behaviors of defenses which cause
this to occur. These behaviors may not perfectly characterize all cases of
masked gradients.%, we find that every defense we study that
%masks gradients exhibits at least one of these behaviors.

\paragraph{One-step attacks perform better than iterative attacks.}
Iterative optimization-based attacks applied in a white-box
setting are strictly
stronger than single-step attacks
and should give strictly superior
performance.
If single-step methods give performance superior to iterative
methods, it is likely that the iterative attack is becoming
stuck in its optimization search at a local minimum.

\paragraph{Black-box attacks are better than white-box attacks.}
The black-box threat model is a strict subset of the white-box threat model,
so attacks in the white-box setting should perform better; if a
defense is obfuscating gradients, then black-box attacks (which do
not use the gradient) often perform better than white-box attacks
\cite{papernot2017blackbox}.

\paragraph{Unbounded attacks do not reach 100\% success.}
With unbounded distortion, any classifier should have
$0\%$ robustness to attack.
If an attack does not reach $100\%$ success with sufficiently large
distortion bound,
this indicates the attack is not performing optimally against the
defense, and the attack should be improved. %For example, with distortion
%$128/255$ any image can be converted to solid grey and model accuracy
%should reach random guessing.

\paragraph{Random sampling finds adversarial examples.}
Brute-force random search (e.g., randomly sampling $10^5$ or more points) within
some $\epsilon$-ball should not find adversarial
examples when gradient-based attacks do not.

\paragraph{Increasing distortion bound does not increase success.}
A larger distortion bound should monotonically increase attack success rate;
significantly increasing distortion bound should result in significantly
higher attack success rate.

\section{Attack Techniques}
\label{sec:techniques}

Generating adversarial examples through optimization-based methods requires
useful gradients obtained through backpropagation~\cite{rumelhart1986backprop}.
Many defenses therefore either intentionally
%``Building Adversary-Resistant Deep Neural Networks without Security through
%Obscurity''
or unintentionally cause gradient descent to fail because of obfuscated
gradients caused by gradient shattering, stochastic gradients, or
vanishing/exploding gradients.
We discuss a number of techniques that we develop to overcome obfuscated gradients.

\subsection{\attacklong{}}

Shattered gradients, caused either unintentionally, e.g. by numerical
instability, or intentionally, e.g. by using non-differentiable operations,
result in nonexistent or incorrect gradients.
To attack defenses where gradients are not readily available, we introduce a
technique we call \attacklong{} (\attack{})~\footnote{The \attack{} approach
can be used on an arbitrary network, even if it is already differentiable, to
obtain a more useful gradient.}.

\subsubsection{A Special Case: \mbox{The Straight-Through Estimator}}

As a special case, we first discuss what amounts to
the straight-through estimator~\cite{bengio2013estimating} applied
to constructing adversarial examples.

Many non-differentiable defenses can be expressed as follows: given a
pre-trained classifier $f(\cdot)$, construct a
preprocessor $g(\cdot)$ and let the secured classifier
$\hat{f}(x) = f(g(x))$
where the preprocessor $g(\cdot)$ satisfies
$g(x) \approx x$ (e.g., such a $g(\cdot)$ may perform
image denoising to remove the adversarial perturbation, as in
\citet{guo2018countering}).
If $g(\cdot)$ is smooth and differentiable, then
computing gradients through the combined network $\hat{f}$
is often sufficient to circumvent the defense~\cite{carlini2017magnet}.
However, recent work has constructed functions
$g(\cdot)$ which are neither smooth nor differentiable, and
therefore can not be backpropagated through to generate
adversarial examples with a white-box attack that requires gradient signal.

Because $g$ is constructed
with the property that $g(x) \approx x$, we can approximate its derivative as
the derivative of the identity function: $\nabla_x g(x) \approx \nabla_x x =
1$. Therefore, we can approximate the derivative of $f(g(x))$ at the point
$\hat{x}$ as:
$$ \left. \nabla_x f(g(x)) \right|_{x = \hat{x}} \approx \left. \nabla_x f(x) \right|_{x = g(\hat{x})} $$
This allows us to compute gradients and therefore
mount a white-box attack.
Conceptually, this attack is simple. We perform forward propagation through the
neural network as usual, but on the backward pass, we replace $g(\cdot)$ with
the identity function. In practice, the implementation can
be expressed in an even simpler way: we approximate $\nabla_x f(g(x))$ by
evaluating $\nabla_x f(x)$ at the point $g(x)$.
This gives us an approximation of the true gradient,
and while not perfect, is sufficiently useful that
when averaged over many iterations of gradient descent
still generates an adversarial example. The math behind the
validity of this approach is similar to the special case.

\subsubsection{Generalized Attack: BPDA}
\label{sec:generalized}

While the above attack is effective for a simple class of networks
expressible as $f(g(x))$ when $g(x) \approx x$, it is not fully general.
We now generalize the above approach into our full attack, which we
call \attacklong{} (BPDA).

Let $f(\cdot) = f^{1 \ldots j}(\cdot)$ be a neural network, and let
$f^{i}(\cdot)$ be a non-differentiable (or not usefully-differentiable)
layer. To approximate $\nabla_x f(x)$,
we first find a differentiable approximation $g(x)$ such that $g(x) \approx
f^i(x)$. Then, we can approximate $\nabla_x f(x)$ by performing the forward
pass through $f(\cdot)$ (and in particular, computing a forward pass through
$f^i(x)$), but on the backward pass, replacing $f^i(x)$ with $g(x)$. Note that
we perform this replacement only on the backward pass. 

As long as the two functions are similar, we find that the slightly inaccurate
gradients still prove useful in constructing an adversarial example. 
Applying BPDA often requires more iterations of gradient descent than without
because each individual gradient descent step is not exactly correct.

We have found applying BPDA is often necessary: replacing $f^i(\cdot)$ with
$g(\cdot)$ on both the forward and backward pass is either completely
ineffective (e.g. with
\citet{song2018pixeldefend}) or many times less effective (e.g. with
\citet{buckman2018thermometer}).

\subsection{Attacking Randomized Classifiers}

Stochastic gradients arise when using randomized transformations to the input
before feeding it to the classifier or when using a stochastic classifier. When
using optimization-based attacks on defenses that employ these techniques, it
is necessary to estimate the gradient of the stochastic function.

\paragraph{Expectation over Transformation.}
For defenses that employ randomized transformations to the input, we apply
Expectation over Transformation (EOT)~\cite{athalye2017synthesizing} to correctly
compute the gradient over the expected transformation to the input.

When attacking a classifier $f(\cdot)$ that first randomly transforms its input
according to a function $t(\cdot)$ sampled from a distribution of
transformations $T$, EOT optimizes the expectation over the
transformation $\mathbb{E}_{t \sim T} f(t(x))$. The optimization problem can be
solved by gradient descent, noting that $\nabla \mathbb{E}_{t \sim T} f(t(x)) =
\mathbb{E}_{t \sim T} \nabla f(t(x))$, differentiating through the classifier
and transformation, and approximating the expectation with samples at each
gradient descent step.

%\paragraph{Stochastic classifiers.}
%For defenses that use stochastic classifiers, we correctly compute the gradient
%by computing gradients over the expectation of random parameters.

\subsection{Reparameterization}

We solve vanishing/exploding gradients by reparameterization.
Assume we are given a classifier $f(g(x))$ where $g(\cdot)$ performs some
optimization loop to transform the input $x$ to a new input $\hat{x}$.
Often times, this optimization loop means that differentiating through
$g(\cdot)$, while
possible, yields exploding or vanishing gradients.

To resolve this, we make a change-of-variable $x=h(z)$ for some function
$h(\cdot)$ such that $g(h(z))=h(z)$
for all $z$, but $h(\cdot)$ is differentiable.
For example, if $g(\cdot)$ projects samples to some manifold in a
specific manner, we might construct $h(z)$ to return points exclusively
on the manifold.
This allows us to compute gradients through $f(h(z))$ and thereby
circumvent the defense.

\section{Case Study: ICLR 2018 Defenses}
\label{sec:breaks}

As a case study for evaluating the prevalence of obfuscated
gradients, we study the ICLR 2018 non-certified defenses that argue
robustness in a white-box threat model.
Each of these defenses argues a high robustness to adaptive,
white-box attacks.
We find that seven of these nine defenses rely on this
phenomenon, and we demonstrate that our
techniques can completely circumvent six of those (and partially
circumvent one) that rely on obfuscated gradients.
We omit two defenses with provable security claims
\cite{raghunathan2018certified,sinha2018certifiable} and
one that only argues black-box security \cite{tramer2018ensemble}.
We include one paper, \citet{ma2018characterizing}, that was
not proposed as a defense \emph{per se}, but suggests a method
to detect adversarial examples.

There is an asymmetry in attacking defenses versus constructing
robust defenses: to show a defense can be
bypassed, it is only necessary to demonstrate one way to
do so; in contrast, a defender must show no attack can succeed.

\newcolumntype{d}[1]{D{\%}{\%}{#1} }
\begin{table}[t]
\small
\centering
\begin{tabular}{p{9em}lld{1}l}
  \toprule
    \textbf{Defense} & Dataset & Distance & \multicolumn{2}{c}{Accuracy} \\
  \midrule
    \citet{buckman2018thermometer} & CIFAR & $0.031$ ($\ell_\infty$) & 0\%* &\\
    \citet{ma2018characterizing} & CIFAR & $0.031$ ($\ell_\infty$) & 5\% &\\
    \citet{guo2018countering} & ImageNet & $0.005$ ($\ell_2$) & 0\%* &\\
    \citet{dhillon2018stochastic} & CIFAR & $0.031$ ($\ell_\infty$) & 0\% &\\
    \citet{xie2018mitigating} & ImageNet & $0.031$ ($\ell_\infty$) & 0\%* &\\
    \citet{song2018pixeldefend} & CIFAR & $0.031$ ($\ell_\infty$) & 9\%* &\\
    \citet{samangouei2018defensegan} & MNIST & $0.005$ ($\ell_2$) & 55\%** &\\
    \midrule
    \citet{madry2018towards} & CIFAR & $0.031$ ($\ell_\infty$) & 47\% &\\
    \citet{na2018cascade} & CIFAR & $0.015$ ($\ell_\infty$) & 15\% &\\
 \bottomrule
\end{tabular}
\vskip 0.1in
\caption{
    \textbf{Summary of Results:} Seven of nine defense techniques accepted at
    ICLR 2018 cause obfuscated gradients and are vulnerable to our attacks.
    Defenses denoted with $*$ propose combining adversarial training; we report
    here the defense alone, see \S\ref{sec:breaks} for full numbers. The
    fundamental principle behind the defense denoted with $**$ has 0\%
    accuracy; in practice, imperfections cause the theoretically optimal attack
    to fail, see \S\ref{sec:defense-gan} for details.
}
  \label{tbl:defenses}
\end{table}

Table~\ref{tbl:defenses} summarizes our results. Of the 9
accepted papers, 7 rely on obfuscated gradients.
Two of these defenses argue robustness on ImageNet, a much harder task
than \cifar{}; and one argues robustness on MNIST, a much easier task than
\cifar{}.
As such, comparing defenses across datasets is difficult.

\subsection{Non-obfuscated Gradients}

\subsubsection{Adversarial Training}

\paragraph{Defense Details.}
Originally proposed by \citet{goodfellow2014explaining},
adversarial training solves a min-max game through
a conceptually simple process: train on adversarial
examples until the model learns to classify them correctly.
Given training data $\mathcal{X}$ and loss function $\ell(\cdot)$,
standard training chooses network weights $\theta$ as
$$ \theta^* = \mathop{\text{arg min}}_\theta\; \mathop{\mathbb{E}}\limits_{(x, y) \in \mathcal{X}} \; \ell(x; y; F_\theta).$$
We study the adversarial training approach of \citet{madry2018towards}
which for a given $\epsilon$-ball solves
$$\theta^* = \mathop{\text{arg min}}_\theta\; \mathop{\mathbb{E}}\limits_{(x, y) \in \mathcal{X}} \left[ \max_{\delta \in [-\epsilon,\epsilon]^N} \ell(x+\delta; y; F_\theta) \right].$$
To approximately solve this formulation, the authors solve the
inner maximization problem by generating adversarial
examples using projected gradient descent.

\paragraph{Discussion.}
We believe this approach does not cause obfuscated gradients: our
experiments with optimization-based attacks do succeed with some
probability (but do not invalidate the claims in the paper).
Further, the authors' evaluation of this defense performs all of the tests for
characteristic behaviors of obfuscated gradients that we list.
However, we note that
(1) adversarial retraining has been shown
to be difficult at ImageNet scale~\cite{kurakin2016scale},
and (2) training exclusively on $\ell_\infty$ adversarial examples provides
only limited robustness to adversarial examples under other
distortion metrics~\cite{sharma2017madry}.

\subsubsection{Cascade Adversarial Training}

Cascade adversarial machine learning \cite{na2018cascade} is closely
related to the above defense. The main difference is that instead of
using iterative methods to generate adversarial examples at each
mini-batch, the authors train a first model, generate adversarial examples
(with iterative methods)
on that model, add these to the training set, and then train a second
model on the augmented dataset only single-step methods for efficiency.
Additionally, the authors construct a ``unified embedding'' and enforce
that the clean and adversarial logits are close under some metric.

\paragraph{Discussion.}
Again, as above, we are unable to reduce the claims made by the authors.
However, these claims are weaker than other defenses (because the authors correctly
performed a strong optimization-based attack~\cite{sp2017:carlini}):
$16\%$ accuracy with $\epsilon=.015$,
compared to over $70\%$ at the same perturbation budget with adversarial
training as in \citet{madry2018towards}.

\subsection{Gradient Shattering}

\subsubsection{Thermometer Encoding}

\paragraph{Defense Details.}
In contrast to prior work \cite{szegedy2013intriguing} which viewed adversarial examples
as ``blind spots'' in neural networks,
\citet{goodfellow2014explaining} argue
that the reason adversarial
examples exist is that neural networks behave in a largely
linear manner.
The purpose of thermometer encoding is to break this linearity.

Given an image $x$, for each pixel color $x_{i,j,c}$, the
$l$-level \emph{thermometer encoding} $\tau(x_{i,j,c})$ is a
$l$-dimensional vector where
$\tau(x_{i,j,c})_k = 1$ if $\text{if}\;\;x_{i,j,c} > k/l$, and $0$ otherwise
%\[ \tau(x_{i,j,c})_k = \begin{cases}
%  1 & \text{if}\;\;x_{i,j,c} > k/l \\
%  0 & \text{otherwise}
%\end{cases}. \]
(e.g., for a 10-level thermometer encoding,
$\tau(0.66) = 1111110000$).
%
%Training networks using thermometer encoding is identical to normal training.

Due to the discrete nature of thermometer encoded values, it is
not possible to directly perform gradient descent on a
thermometer encoded neural network.
The authors therefore construct Logit-Space Projected Gradient
Ascent (LS-PGA) as an attack over the discrete thermometer
encoded inputs.
Using this attack, the authors
perform the adversarial training of \citet{madry2018towards}
on thermometer encoded networks.

On \cifar{}, just performing thermometer encoding was found to give
$50\%$ accuracy within $\epsilon=0.031$ under $\ell_\infty$ distortion.
By performing adversarial training with $7$ steps of LS-PGA, robustness
increased to $80\%$.

\paragraph{Discussion.}
While the intention behind this defense is to break
the local linearity of neural networks, we find that this
defense in fact causes gradient shattering.
This can be observed through their black-box attack evaluation:
adversarial examples generated on a standard adversarially trained
model transfer to a thermometer encoded model reducing the accuracy
to $67\%$, well below the $80\%$ robustness to the white-box
iterative attack.

\paragraph{Evaluation.}
We use the \attack{} approach from \S\ref{sec:generalized},
where we let $f(x)=\tau(x)$.
Observe that if we define
\vspace{-.5em}
\[ \hat\tau(x_{i,j,c})_k = \min(\max(x_{i,j,c} - k/l, 0), 1) \]
then
\vspace{-.5em}
\[ \tau(x_{i,j,c})_k = \text{floor}(\hat\tau(x_{i,j,c})_k) \]
so we can let $g(x)=\hat\tau(x)$ and replace the backwards
pass with the function $g(\cdot)$.

LS-PGA only reduces model accuracy to $50\%$ on a
thermometer-encoded model trained \emph{without} adversarial training
(bounded by $\epsilon=0.031$).
In contrast, we achieve $1\%$ model accuracy with the lower $\epsilon=0.015$
(and $0\%$ with $\epsilon=0.031$).
This shows no measurable improvement from standard models, trained
without thermometer encoding.

When we attack a thermometer-encoded adversarially trained model~\footnote{That is,
  a thermometer encoded model that is trained using the approach of \cite{madry2018towards}.},
we are able to reproduce the $80\%$ accuracy at $\epsilon=0.031$
claim against LS-PGA. However, our attack reduces model accuracy
to $30\%$.
This is significantly \emph{weaker} than the original \citet{madry2018towards}
model that does not use thermometer encoding.
Because this model is trained against the (comparatively weak)
LS-PGA attack, it is unable to adapt to the stronger attack we
present above.
\ifdefined\isarxiv
Figure~\ref{fig:thermometer} shows a comparison of thermometer encoding, with
and without adversarial training, against the baseline classifier, over a range
of perturbation magnitudes, demonstrating that thermometer encoding provides
limited value.
\fi

\ifdefined\isarxiv
\begin{figure}
    \center
    \includegraphics[scale=.75]{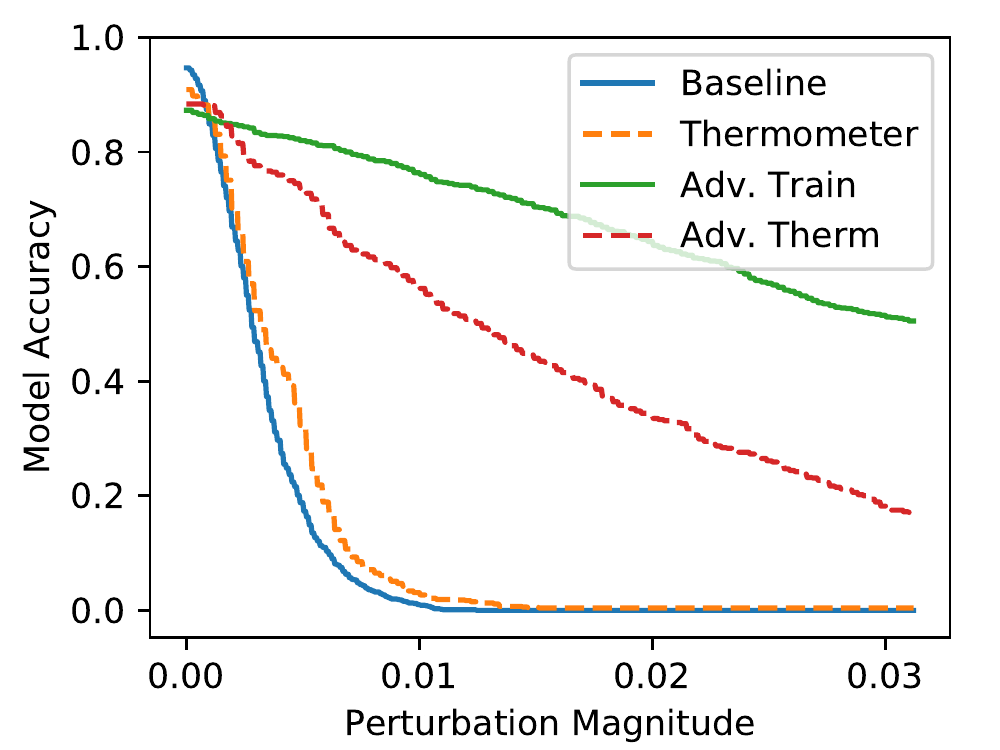}
    \caption{Model accuracy versus distortion (under $\ell_\infty$).
    Adversarial training increases robustness to $50\%$ at $\epsilon=0.031$;
    thermometer encoding by itself provides limited value, and when coupled
    with adversarial training performs worse than adversarial training alone.}
    \label{fig:thermometer}
\end{figure}
\fi

\subsubsection{Input Transformations}

\paragraph{Defense Details.}
\citet{guo2018countering} propose five input transformations to counter
adversarial examples.

As a baseline, the authors evaluate \emph{image cropping and rescaling},
\emph{bit-depth reduction}, and \emph{JPEG compression}.
Then the authors suggest two new transformations:
(a) randomly drop pixels and restore them
by performing \emph{total variance minimization}; and (b)
\emph{image quilting}: reconstruct images by replacing small
patches with patches from ``clean'' images, using minimum graph cuts in overlapping
boundary regions to remove edge artifacts.

%% \begin{itemize}
%%   \setlength\itemsep{-.2em}
%%     \item Perform random \textit{image cropping and rescaling}, averaging
%%       over multiple runs.

%%     \item Quantize images with \textit{bit-depth reduction.}

%%     \item Apply \textit{JPEG compression} to remove perturbations.

%%     \item Randomly drop pixels, and restore them
%%       by performing \textit{total variance minimization}.

%%     \item \textit{Image quilting:} Reconstruct images by replacing all $5\times5$
%%       patches with
%%         patches from ``clean'' images, using minimum graph cuts in overlapping
%%         boundary regions to remove edge artifacts.
%% \end{itemize}

The authors explore different combinations of input transformations along with
different underlying ImageNet classifiers, including adversarially trained
models.
They find that input transformations provide protection even with a vanilla
classifier.

\paragraph{Discussion.}

The authors find that a ResNet-50 classifier provides a varying degree of
accuracy for each of the five proposed input transformations
under the
strongest attack with a normalized $\ell_2$ dissimilarity of $0.01$, with the
strongest defenses achieving over $60\%$ top-1 accuracy. We reproduce these results
when evaluating an InceptionV3 classifier.

The authors do not succeed in white-box attacks, crediting lack of access to
test-time randomness as ``particularly crucial in developing strong
defenses''~\cite{guo2018countering}.~\footnote{This defense may be stronger in
  a threat model where the adversary does not have complete information about
  the exact quilting process used (personal communication with authors).}

\paragraph{Evaluation.}

It is possible to bypass each defense independently (and ensembles of defenses
usually are not much stronger than the strongest sub-component~\cite{he2017adversarial}).
We circumvent image cropping and rescaling with a direct application of
EOT.
To circumvent bit-depth reduction and JPEG compression, we use \attack{} and
approximate the backward pass with the identity function.
To circumvent total variance minimization and image quilting, which are both
non-differentiable and randomized, we apply EOT and use \attack{} to
approximate the gradient through the transformation.
With our attack, we achieve $100\%$ targeted attack success rate
and accuracy drops to $0\%$ for the strongest defense under the smallest
perturbation budget considered in \citet{guo2018countering}, a root-mean-square
perturbation of $0.05$ (and a ``normalized'' $\ell_2$ perturbation as defined
in \citet{guo2018countering} of $0.01$).

\subsubsection{Local Intrinsic Dimensionality (LID)}

LID is a general-purpose metric that measures the distance from an input
to its neighbors.
\citet{ma2018characterizing} propose using LID to characterize
properties of adversarial examples.
The authors emphasize that this classifier \emph{is not intended as a defense}
against adversarial examples~\footnote{Personal communication with authors.},
however the authors argue that it is a robust method for detecting
adversarial examples that is not easy to evade by attempting their own
adaptive attack and showing it fails.

\paragraph{Analysis Overview.}
Instead of actively attacking the
detection method,
we find that LID is not able to
detect high confidence adversarial examples~\cite{carlini2017adversarial},
even in the unrealistic threat model where the adversary is
\emph{entirely oblivious} to the defense and generates adversarial examples
on the original classifier.
A full discussion of this attack is given in \supplement{}~\ref{app:lid}.

\subsection{Stochastic Gradients}

\subsubsection{Stochastic Activation Pruning (SAP)}

\paragraph{Defense Details.}
SAP \cite{dhillon2018stochastic} introduces randomness into the evaluation of a neural network
to defend against adversarial examples.
SAP randomly drops some neurons of each layer $f^i$ to 0 with
probability proportional to their absolute value.
That is, SAP essentially applies dropout at each layer where instead of
dropping with uniform probability, nodes are dropped with a weighted
distribution.
Values which are retained are scaled up (as is done in dropout) to
retain accuracy.
Applying SAP decreases clean classification accuracy slightly, with a
higher drop probability decreasing accuracy, but increasing robustness.
We study various levels of drop probability and find they lead to similar
robustness numbers.
%% %
%% For the hidden activation vector $h^i = f^{1..i}(x)$
%% at layer $i$, SAP defines a probability distribution
%% \[ p^i_j = |h^i_j| \cdot \left(\sum\limits_{k=1}^m h^i_m\right)^{-1}. \]
%% That is, $p^i_j$ is proportional to the magnitude of $h^i_j$ compared
%% to the magnitude of the other neurons at this layer.
%% SAP then computes a modified distribution
%% \[ q^i_j = {1 - (1 - p^i_j)^{r}} \]
%% where $r$ is a defense hyperparameter (discussed below).

%% Then, each $h^i_j$ is updated to a new value $\hat{h}^i_j$
%% by dropping it with probability $q^i_j$ and keeping it otherwise
%% \[ \hat{h}^i_j = \begin{cases}
%%   {h^i_j \over q^i_j} & \text{with probability} \;\;\; q^i_j \\
%%   0 & \text{with probability} \;\;\; 1 - q^i_j \\
%% \end{cases}
%% \]
%% so that values less likely to be sampled are scaled up accordingly.
%% %
%% The value $r$ is chosen to keep a large enough fraction of the
%% neurons that the accuracy remains high, but not so large that all
%% neurons are kept.
%% %
%% We follow the authors advice and choose $r$ so the test accuracy
%% drops by only $5\%$.

\paragraph{Discussion.}
The authors only evaluate SAP by taking a single step
in the gradient direction~\cite{dhillon2018stochastic}.
While taking a single step in the direction of the gradient can be
effective on non-randomized neural networks, when randomization is
used, computing the gradient with respect to one sample of the
randomness is ineffective.
%
%To see this, we observe that an adversarial example constructed
%in this way is often \emph{no longer} adversarial with fresh
%randomness.

\paragraph{Evaluation.}
To resolve this difficulty,
we estimate the gradients by computing the expectation over
instantiations of randomness. At each iteration of gradient
descent, instead of taking a step in the direction of $\nabla_x f(x)$
we move in the direction of
$\sum_{i=1}^k \nabla_x f(x)$
where each invocation is randomized with SAP.
We have found that choosing $k=10$ provides useful gradients.
We additionally had to resolve a numerical instability when computing
gradients: this defense caused computing a backward pass to cause
exploding gradients due to division by numbers very close to 0.
%
%We resolve this by clipping gradients or through stable
%numerical techniques.

With these approaches,
we are able to reduce SAP model accuracy to $9\%$ at $\epsilon=.015$, and
$0\%$ at $\epsilon=0.031$.
If we consider an attack successful only when an
example is classified incorrectly $10$ times out of $10$ (and consider
it correctly classified if it is ever classified as the correct label),
model accuracy is below $10\%$ with $\epsilon=0.031$.

\subsubsection{Mitigating through Randomization}

\paragraph{Defense Details.}
\citet{xie2018mitigating} propose to defend against adversarial examples by adding a
randomization layer before the input to the classifier. For a classifier that
takes a $299 \times 299$ input, the defense first randomly rescales the image
to a $r \times r$ image, with $r \in [299, 331)$, and then randomly
zero-pads the image so that the result is $331 \times 331$. The output is then
fed to the classifier. %With these parameters for the randomization layer, there
%are $12528$ different randomizations for any given input.

\paragraph{Discussion.} The authors consider three attack scenarios: vanilla
attack (an attack on the original classifier), single-pattern attack (an attack
assuming some fixed randomization pattern), and ensemble-pattern attack (an
attack over a small ensemble of fixed randomization patterns). The authors
strongest attack reduces InceptionV3 model accuracy to $32.8\%$ top-1
accuracy (over images that were originally classified correctly).

The authors dismiss a stronger attack
over larger choices of randomness,
stating that it would be ``computationally \emph{impossible}'' (emphasis ours)
and that such an attack ``may not even converge'' \cite{xie2018mitigating}.
%``the strongest attack would be that attackers consider ALL possible
%patterns of the defense models when generating the adversarial examples.
%However, this is computationally impossible, because failing a large number of
%patterns (e.g. 12528 here) at the same time takes extremely long time, and may
%not even converge''

\paragraph{Evaluation.}
We find the authors' ensemble attack overfits to the ensemble with fixed
randomization.
We bypass this defense by applying EOT, optimizing over the (in this
case, discrete) distribution of transformations.

Using this attack, even if we consider the attack
successful only when an example is classified incorrectly $10$ times out of
$10$, we achieve $100\%$ targeted attack success rate and reduce the
accuracy of the classifier from $32.8\%$ to $0.0\%$ with a maximum
$\ell_\infty$ perturbation of $\epsilon = 0.031$.

\subsection{Vanishing \& Exploding Gradients}

\subsubsection{PixelDefend}

\paragraph{Defense Details.}
\citet{song2018pixeldefend} propose using a PixelCNN generative model to project
a potential adversarial example back onto the data manifold before feeding it
into a classifier.
The authors argue that adversarial examples mainly lie in the low-probability
region of the data distribution.
PixelDefend ``purifies'' adversarially perturbed images prior to classification
by using a greedy decoding procedure to approximate finding the highest
probability example within an $\epsilon$-ball of the input image.

\paragraph{Discussion.}

The authors evaluate PixelDefend on \cifar{} over various
classifiers and perturbation budgets.
With a maximum $\ell_\infty$ perturbation of $\epsilon = 0.031$, PixelDefend
claims $46\%$ accuracy (with a vanilla ResNet classifier).
The authors dismiss the possibility of end-to-end
attacks on PixelDefend due to the difficulty of differentiating through an
unrolled version of PixelDefend due to vanishing gradients and computation
cost.
%
%Furthermore, they claim that because the generative model and classifier
%are trained separately and have independent parameters, the perturbation
%direction that leads to higher probability images has smaller correlation with
%the perturbation direction that results in misclassification.
%\todo{unfortunately, it's broken due to xx}

\paragraph{Evaluation.}

We sidestep the problem of computing gradients through an unrolled version of
PixelDefend by approximating gradients with \attack{}, and we successfully mount an
end-to-end attack using this technique~\footnote{In place of a PixelCNN, due to
the availability of a pre-trained model, we use a
PixelCNN++~\cite{Salimans2017PixeCNN} and discretize the mixture of logistics
to produce a 256-way softmax. %Due to compute
%limitations, we evaluate our attack over a random sample of 500 images from the
%test set.
}. With this attack, we can reduce the accuracy of a naturally trained
classifier which achieves $95\%$ accuracy to $9\%$ with a maximum $\ell_\infty$
perturbation of $\epsilon = 0.031$. We find that combining
adversarial training~\cite{madry2018towards} with PixelDefend provides no additional
robustness over just using the adversarially trained classifier.

\subsubsection{Defense-GAN}
\label{sec:defense-gan}
Defense-GAN \cite{samangouei2018defensegan} uses a
Generative Adversarial Network \cite{goodfellow2014generative}
to project
samples onto the manifold of the generator before classifying
them.
That is, the intuition behind this defense is nearly identical to
PixelDefend, but using a GAN instead of a PixelCNN. We therefore
summarize results here
and present the full details in \supplement{}~\ref{app:dgan}.

\paragraph{Analysis Overview.}
Defense-GAN is not argued secure on CIFAR-10, so we use MNIST.
We find that adversarial examples exist on the manifold
defined by the generator. That is, we show that we are able to construct
an adversarial example $x'=G(z)$ so that $x' \approx x$ but $c(x) \ne c(x')$.
As such, a perfect projector would not modify this example $x'$ because
it exists on the manifold described by the generator.
However, while this attack would defeat a perfect projector
mapping $x$ to its nearest point on $G(z)$, the imperfect
gradient descent based approach taken by Defense-GAN does
not perfectly preserve points on the manifold.
We therefore construct a second attack using BPDA to evade Defense-GAN,
although at only a $45\%$ success rate.

\section{Discussion}
\label{sec:discussion}

Having demonstrated attacks on these seven defenses, we now take a step back
and discuss the method of evaluating a defense against adversarial examples.

The papers we study use a variety of approaches in evaluating robustness of the
proposed defenses. We list what we believe to be the most
important points to keep in mind while building and evaluating defenses. Much
of what we describe below has been discussed in prior
work~\cite{carlini2017adversarial,madry2018towards}; we repeat these points
here and offer our own perspective for completeness.

\subsection{Define a (realistic) threat model}

A threat model specifies the conditions under which a defense argues
security: a precise threat model allows for an exact understanding of the
setting under which the defense is meant to work. Prior work has used words
including \emph{white-box}, \emph{grey-box}, \emph{black-box}, and \emph{no-box} to
describe slightly different threat models, often overloading the same word.

Instead of attempting to, yet again, redefine the vocabulary, we enumerate the
various aspects of a defense that might be revealed to the adversary or held
secret to the defender:
\emph{model architecture} and \emph{model weights};
\emph{training algorithm} and \emph{training data};
test time \emph{randomness} (either the values chosen or the distribution);
and, if the model weights are held secret, whether \emph{query access} is
allowed (and if so, the type of output, e.g. logits or only the top label).

While there are some aspects of a defense that might be held secret,
threat models should not contain unrealistic constraints.
We believe any compelling threat model should at the very least
grant knowledge of the model architecture, training algorithm,
and allow query access.

It is not meaningful to restrict the
computational power of an adversary artificially (e.g., to fewer
than several thousand attack iterations).
If two defenses are equally
robust but generating adversarial examples on one takes
one second and another takes ten seconds, the robustness \emph{has not
  increased}.
%
%For some defenses we study, increasing the computation time
%(from one second per adversarial example to five seconds per
%adversarial example) increases attack success by over a factor
%of two.

\subsection{Make specific, testable claims}

Specific, testable claims in a clear threat model precisely convey the claimed
robustness of a defense.
For example, a complete claim might be: ``We achieve
$90\%$ accuracy when bounded by $\ell_\infty$ distortion
with $\epsilon=0.031$, when the attacker has full
white-box access.''%, or might claim that mean
%$\ell_2$ distortion to adversarial examples increases by a factor of two from a
%baseline model to a secured model (in which case, the baseline should also be
%clearly defined).

In this paper, we study all papers under the threat model the authors define.
However, if a paper is evaluated under a \emph{different} threat model,
explicitly stating so makes it clear that the original paper's claims are not
being violated.

%A defense can never achieve complete robustness against \emph{unbounded}
%attacks: with unlimited distortion any image can be converted into any other,
%yielding  $100\%$ ``success''.

A defense being specified completely, with all hyperparameters given, is a
prerequisite for claims to be testable.
Releasing source code and a pre-trained model along with the paper describing a
specific threat model and robustness claims is perhaps the most useful method
of making testable claims.
At the time of writing this paper, four of the defenses we study made complete source code
available~\cite{madry2018towards,ma2018characterizing,guo2018countering,xie2018mitigating}.

\subsection{Evaluate against adaptive attacks}

A strong defense is robust \emph{not only} against existing attacks,
but \emph{also} against
future attacks within the specified threat model. %While certified
%defenses succeed in reasoning about all possible attacks, it is often
%challenging to do so.
%However, actively evaluating a defense with new
%defense-aware attacks crafted specifically to circumvent the defense helps
%justify claims of security.
A necessary component of any defense proposal is therefore an attempt at
an adaptive attack.

An \emph{adaptive attack} is one that is constructed \emph{after} a defense has been
completely specified, where the adversary takes advantage of knowledge of the
defense and is only restricted by the threat model.
%
%If a defense can be circumvented by an adaptive attack, giving a way to prevent
%that specific attack (e.g. by tweaking a hyperparameter) does not imply
%robustness.
One useful attack approach is to perform many attacks and report the mean over the
best attack
\emph{per image}. That is, for a set of attacks $a \in \mathcal{A}$
instead of reporting the value
$\min\limits_{a \in \mathcal{A}} \mathop{\text{mean}}\limits_{x \in \mathcal{A}} f(a(x))$
report
$\mathop{\text{mean}}\limits_{x \in \mathcal{A}} \min\limits_{a \in \mathcal{A}} f(a(x))$.

If a defense is modified after an evaluation, an adaptive attack is one that
considers knowledge of the \emph{new} defense.
In this way, concluding an evaluation with a final adaptive attack can be seen
as analogous to evaluating a model on the test data.

\section{Conclusion}
\label{sec:conclusion}

Constructing defenses to adversarial examples requires defending
against not only existing attacks but also future attacks that
may be developed.
In this paper, we identify \emph{obfuscated gradients}, a phenomenon exhibited
by certain defenses that makes standard gradient-based methods fail to generate
adversarial examples.
We develop three attack techniques to bypass three different types of
obfuscated gradients.
To evaluate the applicability of our techniques, we use the
ICLR 2018 defenses as a case study, circumventing seven of
nine accepted defenses.

More generally, we hope that future work will be able to avoid
relying on obfuscated gradients (and other methods that only
prevent gradient descent-based attacks) for perceived robustness,
and use our evaluation approach to
detect when this occurs.
Defending against adversarial examples is an important area
of research and we believe performing a careful, thorough evaluation is a
critical step that can not be overlooked when designing defenses.

\ifdefined\isarxiv
\else
\ifnum\value{page}>8
\todo{paper is too long: limit is 8 pages for main content}
\else
\fi
\fi

\section*{Acknowledgements}

We are grateful to Aleksander Madry, Andrew Ilyas, and Aditi Raghunathan for
helpful comments on an early draft of this paper.
We thank Bo Li, Xingjun Ma, Laurens van der Maaten, Aurko Roy,
Yang Song, and Cihang Xie
for useful discussion and insights on their defenses.

This work was partially supported by the National Science
Foundation through award CNS-1514457, Qualcomm,
and the Hewlett Foundation through the Center for Long-Term
Cybersecurity.

\bibliography{paper}

\begin{thebibliography}{34}
\providecommand{\natexlab}[1]{#1}
\providecommand{\url}[1]{\texttt{#1}}
\expandafter\ifx\csname urlstyle\endcsname\relax
  \providecommand{\doi}[1]{doi: #1}\else
  \providecommand{\doi}{doi: \begingroup \urlstyle{rm}\Url}\fi

\bibitem[Amsaleg et~al.(2015)Amsaleg, Chelly, Furon, Girard, Houle,
  Kawarabayashi, and Nett]{amsaleg2015estimating}
Amsaleg, L., Chelly, O., Furon, T., Girard, S., Houle, M.~E., Kawarabayashi,
  K.-i., and Nett, M.
\newblock Estimating local intrinsic dimensionality.
\newblock In \emph{Proceedings of the 21th ACM SIGKDD International Conference
  on Knowledge Discovery and Data Mining}, pp.\  29--38. ACM, 2015.

\bibitem[Athalye et~al.(2017)Athalye, Engstrom, Ilyas, and
  Kwok]{athalye2017synthesizing}
Athalye, A., Engstrom, L., Ilyas, A., and Kwok, K.
\newblock Synthesizing robust adversarial examples.
\newblock \emph{arXiv preprint arXiv:1707.07397}, 2017.

\bibitem[Bengio et~al.(2013)Bengio, L{\'e}onard, and
  Courville]{bengio2013estimating}
Bengio, Y., L{\'e}onard, N., and Courville, A.
\newblock Estimating or propagating gradients through stochastic neurons for
  conditional computation.
\newblock \emph{arXiv preprint arXiv:1308.3432}, 2013.

\bibitem[Biggio et~al.(2013)Biggio, Corona, Maiorca, Nelson, {\v{S}}rndi{\'c},
  Laskov, Giacinto, and Roli]{biggio2013evasion}
Biggio, B., Corona, I., Maiorca, D., Nelson, B., {\v{S}}rndi{\'c}, N., Laskov,
  P., Giacinto, G., and Roli, F.
\newblock Evasion attacks against machine learning at test time.
\newblock In \emph{Joint European Conference on Machine Learning and Knowledge
  Discovery in Databases}, pp.\  387--402. Springer, 2013.

\bibitem[Buckman et~al.(2018)Buckman, Roy, Raffel, and
  Goodfellow]{buckman2018thermometer}
Buckman, J., Roy, A., Raffel, C., and Goodfellow, I.
\newblock Thermometer encoding: One hot way to resist adversarial examples.
\newblock \emph{International Conference on Learning Representations}, 2018.
\newblock URL \url{https://openreview.net/forum?id=S18Su--CW}.
\newblock accepted as poster.

\bibitem[Carlini \& Wagner(2017{\natexlab{a}})Carlini and
  Wagner]{carlini2017adversarial}
Carlini, N. and Wagner, D.
\newblock Adversarial examples are not easily detected: Bypassing ten detection
  methods.
\newblock \emph{AISec}, 2017{\natexlab{a}}.

\bibitem[Carlini \& Wagner(2017{\natexlab{b}})Carlini and
  Wagner]{carlini2017magnet}
Carlini, N. and Wagner, D.
\newblock Magnet and ``efficient defenses against adversarial attacks'' are not
  robust to adversarial examples.
\newblock \emph{arXiv preprint arXiv:1711.08478}, 2017{\natexlab{b}}.

\bibitem[Carlini \& Wagner(2017{\natexlab{c}})Carlini and
  Wagner]{sp2017:carlini}
Carlini, N. and Wagner, D.
\newblock Towards evaluating the robustness of neural networks.
\newblock In \emph{IEEE Symposium on Security \& Privacy}, 2017{\natexlab{c}}.

\bibitem[Dhillon et~al.(2018)Dhillon, Azizzadenesheli, Bernstein, Kossaifi,
  Khanna, Lipton, and Anandkumar]{dhillon2018stochastic}
Dhillon, G.~S., Azizzadenesheli, K., Bernstein, J.~D., Kossaifi, J., Khanna,
  A., Lipton, Z.~C., and Anandkumar, A.
\newblock Stochastic activation pruning for robust adversarial defense.
\newblock \emph{International Conference on Learning Representations}, 2018.
\newblock URL \url{https://openreview.net/forum?id=H1uR4GZRZ}.
\newblock accepted as poster.

\bibitem[Goodfellow et~al.(2014{\natexlab{a}})Goodfellow, Pouget-Abadie, Mirza,
  Xu, Warde-Farley, Ozair, Courville, and Bengio]{goodfellow2014generative}
Goodfellow, I., Pouget-Abadie, J., Mirza, M., Xu, B., Warde-Farley, D., Ozair,
  S., Courville, A., and Bengio, Y.
\newblock Generative adversarial nets.
\newblock In \emph{Advances in neural information processing systems}, pp.\
  2672--2680, 2014{\natexlab{a}}.

\bibitem[Goodfellow et~al.(2014{\natexlab{b}})Goodfellow, Shlens, and
  Szegedy]{goodfellow2014explaining}
Goodfellow, I.~J., Shlens, J., and Szegedy, C.
\newblock Explaining and harnessing adversarial examples.
\newblock \emph{arXiv preprint arXiv:1412.6572}, 2014{\natexlab{b}}.

\bibitem[Gulrajani et~al.(2017)Gulrajani, Ahmed, Arjovsky, Dumoulin, and
  Courville]{gulrajani2017improved}
Gulrajani, I., Ahmed, F., Arjovsky, M., Dumoulin, V., and Courville, A.
\newblock Improved training of wasserstein gans.
\newblock \emph{arXiv preprint arXiv:1704.00028}, 2017.

\bibitem[Guo et~al.(2018)Guo, Rana, Cisse, and van~der
  Maaten]{guo2018countering}
Guo, C., Rana, M., Cisse, M., and van~der Maaten, L.
\newblock Countering adversarial images using input transformations.
\newblock \emph{International Conference on Learning Representations}, 2018.
\newblock URL \url{https://openreview.net/forum?id=SyJ7ClWCb}.
\newblock accepted as poster.

\bibitem[He et~al.(2016)He, Zhang, Ren, and Sun]{he2016deep}
He, K., Zhang, X., Ren, S., and Sun, J.
\newblock Deep residual learning for image recognition.
\newblock In \emph{Proceedings of the IEEE conference on computer vision and
  pattern recognition}, pp.\  770--778, 2016.

\bibitem[He et~al.(2017)He, Wei, Chen, Carlini, and Song]{he2017adversarial}
He, W., Wei, J., Chen, X., Carlini, N., and Song, D.
\newblock Adversarial example defenses: Ensembles of weak defenses are not
  strong.
\newblock \emph{arXiv preprint arXiv:1706.04701}, 2017.

\bibitem[Ilyas et~al.(2017)Ilyas, Jalal, Asteri, Daskalakis, and
  Dimakis]{ilyas2017robust}
Ilyas, A., Jalal, A., Asteri, E., Daskalakis, C., and Dimakis, A.~G.
\newblock The robust manifold defense: Adversarial training using generative
  models.
\newblock \emph{arXiv preprint arXiv:1712.09196}, 2017.

\bibitem[Kurakin et~al.(2016{\natexlab{a}})Kurakin, Goodfellow, and
  Bengio]{kurakin2016adversarial}
Kurakin, A., Goodfellow, I., and Bengio, S.
\newblock Adversarial examples in the physical world.
\newblock \emph{arXiv preprint arXiv:1607.02533}, 2016{\natexlab{a}}.

\bibitem[Kurakin et~al.(2016{\natexlab{b}})Kurakin, Goodfellow, and
  Bengio]{kurakin2016scale}
Kurakin, A., Goodfellow, I.~J., and Bengio, S.
\newblock Adversarial machine learning at scale.
\newblock \emph{arXiv preprint arXiv:1611.01236}, 2016{\natexlab{b}}.

\bibitem[Ma et~al.(2018)Ma, Li, Wang, Erfani, Wijewickrema, Schoenebeck, Houle,
  Song, and Bailey]{ma2018characterizing}
Ma, X., Li, B., Wang, Y., Erfani, S.~M., Wijewickrema, S., Schoenebeck, G.,
  Houle, M.~E., Song, D., and Bailey, J.
\newblock Characterizing adversarial subspaces using local intrinsic
  dimensionality.
\newblock \emph{International Conference on Learning Representations}, 2018.
\newblock URL \url{https://openreview.net/forum?id=B1gJ1L2aW}.
\newblock accepted as oral presentation.

\bibitem[Madry et~al.(2018)Madry, Makelov, Schmidt, Tsipras, and
  Vladu]{madry2018towards}
Madry, A., Makelov, A., Schmidt, L., Tsipras, D., and Vladu, A.
\newblock Towards deep learning models resistant to adversarial attacks.
\newblock \emph{International Conference on Learning Representations}, 2018.
\newblock URL \url{https://openreview.net/forum?id=rJzIBfZAb}.
\newblock accepted as poster.

\bibitem[Na et~al.(2018)Na, Ko, and Mukhopadhyay]{na2018cascade}
Na, T., Ko, J.~H., and Mukhopadhyay, S.
\newblock Cascade adversarial machine learning regularized with a unified
  embedding.
\newblock In \emph{International Conference on Learning Representations}, 2018.
\newblock URL \url{https://openreview.net/forum?id=HyRVBzap-}.

\bibitem[Papernot et~al.(2017)Papernot, McDaniel, Goodfellow, Jha, Celik, and
  Swami]{papernot2017blackbox}
Papernot, N., McDaniel, P., Goodfellow, I., Jha, S., Celik, Z.~B., and Swami,
  A.
\newblock Practical black-box attacks against machine learning.
\newblock In \emph{Proceedings of the 2017 ACM on Asia Conference on Computer
  and Communications Security}, ASIA CCS '17, pp.\  506--519, New York, NY,
  USA, 2017. ACM.
\newblock ISBN 978-1-4503-4944-4.
\newblock \doi{10.1145/3052973.3053009}.
\newblock URL \url{http://doi.acm.org/10.1145/3052973.3053009}.

\bibitem[Raghunathan et~al.(2018)Raghunathan, Steinhardt, and
  Liang]{raghunathan2018certified}
Raghunathan, A., Steinhardt, J., and Liang, P.
\newblock Certified defenses against adversarial examples.
\newblock \emph{International Conference on Learning Representations}, 2018.
\newblock URL \url{https://openreview.net/forum?id=Bys4ob-Rb}.

\bibitem[Rumelhart et~al.(1986)Rumelhart, Hinton, and
  Williams]{rumelhart1986backprop}
Rumelhart, D.~E., Hinton, G.~E., and Williams, R.~J.
\newblock Learning representations by back-propagating errors.
\newblock \emph{Nature}, 323:\penalty0 533--536, 1986.

\bibitem[Salimans et~al.(2017)Salimans, Karpathy, Chen, and
  Kingma]{Salimans2017PixeCNN}
Salimans, T., Karpathy, A., Chen, X., and Kingma, D.~P.
\newblock Pixelcnn++: A pixelcnn implementation with discretized logistic
  mixture likelihood and other modifications.
\newblock In \emph{ICLR}, 2017.

\bibitem[Samangouei et~al.(2018)Samangouei, Kabkab, and
  Chellappa]{samangouei2018defensegan}
Samangouei, P., Kabkab, M., and Chellappa, R.
\newblock Defense-gan: Protecting classifiers against adversarial attacks using
  generative models.
\newblock \emph{International Conference on Learning Representations}, 2018.
\newblock URL \url{https://openreview.net/forum?id=BkJ3ibb0-}.
\newblock accepted as poster.

\bibitem[Sharma \& Chen(2017)Sharma and Chen]{sharma2017madry}
Sharma, Y. and Chen, P.-Y.
\newblock Attacking the madry defense model with ${L}_1$-based adversarial
  examples.
\newblock \emph{arXiv preprint arXiv:1710.10733}, 2017.

\bibitem[Sinha et~al.(2018)Sinha, Namkoong, and Duchi]{sinha2018certifiable}
Sinha, A., Namkoong, H., and Duchi, J.
\newblock Certifiable distributional robustness with principled adversarial
  training.
\newblock \emph{International Conference on Learning Representations}, 2018.
\newblock URL \url{https://openreview.net/forum?id=Hk6kPgZA-}.

\bibitem[Song et~al.(2018)Song, Kim, Nowozin, Ermon, and
  Kushman]{song2018pixeldefend}
Song, Y., Kim, T., Nowozin, S., Ermon, S., and Kushman, N.
\newblock Pixeldefend: Leveraging generative models to understand and defend
  against adversarial examples.
\newblock \emph{International Conference on Learning Representations}, 2018.
\newblock URL \url{https://openreview.net/forum?id=rJUYGxbCW}.
\newblock accepted as poster.

\bibitem[Szegedy et~al.(2013)Szegedy, Zaremba, Sutskever, Bruna, Erhan,
  Goodfellow, and Fergus]{szegedy2013intriguing}
Szegedy, C., Zaremba, W., Sutskever, I., Bruna, J., Erhan, D., Goodfellow, I.,
  and Fergus, R.
\newblock Intriguing properties of neural networks.
\newblock \emph{ICLR}, 2013.

\bibitem[Szegedy et~al.(2016)Szegedy, Vanhoucke, Ioffe, Shlens, and
  Wojna]{szegedy2016rethinking}
Szegedy, C., Vanhoucke, V., Ioffe, S., Shlens, J., and Wojna, Z.
\newblock Rethinking the inception architecture for computer vision.
\newblock In \emph{Proceedings of the IEEE Conference on Computer Vision and
  Pattern Recognition}, pp.\  2818--2826, 2016.

\bibitem[Tram{\`e}r et~al.(2018)Tram{\`e}r, Kurakin, Papernot, Goodfellow,
  Boneh, and McDaniel]{tramer2018ensemble}
Tram{\`e}r, F., Kurakin, A., Papernot, N., Goodfellow, I., Boneh, D., and
  McDaniel, P.
\newblock Ensemble adversarial training: Attacks and defenses.
\newblock \emph{International Conference on Learning Representations}, 2018.
\newblock URL \url{https://openreview.net/forum?id=rkZvSe-RZ}.
\newblock accepted as poster.

\bibitem[Xie et~al.(2018)Xie, Wang, Zhang, Ren, and Yuille]{xie2018mitigating}
Xie, C., Wang, J., Zhang, Z., Ren, Z., and Yuille, A.
\newblock Mitigating adversarial effects through randomization.
\newblock \emph{International Conference on Learning Representations}, 2018.
\newblock URL \url{https://openreview.net/forum?id=Sk9yuql0Z}.
\newblock accepted as poster.

\bibitem[Zagoruyko \& Komodakis(2016)Zagoruyko and
  Komodakis]{zagoruyko2016wide}
Zagoruyko, S. and Komodakis, N.
\newblock Wide residual networks.
\newblock \emph{arXiv preprint arXiv:1605.07146}, 2016.

\end{thebibliography}
\bibliographystyle{icml2018}

\clearpage
\appendix
\section{Local Intrinsic Dimensionality}
\label{app:lid}

\paragraph{Defense Details.}
The Local Intrinsic Dimensionality \cite{amsaleg2015estimating}
``assesses the
space-filling capability of the region surrounding a reference
example, based on
the distance distribution of the example to its neighbors'' \cite{ma2018characterizing}.
The authors present evidence that the LID is significantly larger for adversarial examples
generated by existing attacks than for normal images, and they construct
a classifier that can distinguish these adversarial images from normal images.
Again, the authors indicate that LID \emph{is not intended as a defense} and only
should be used to explore properties of adversarial examples.
However, it would be natural to wonder whether it would be effective as
a defense, so we study its robustness; our results confirm that it is
not adequate as a defense.
The method used to compute the LID relies on finding the
$k$ nearest neighbors, a non-differentiable
operation, rendering gradient descent based methods ineffective.

Let $\mathcal{S}$ be a mini-batch of $N$ clean examples.
Let $r_i(x)$ denote the distance (under metric $d(x,y)$)
between sample $x$ and its $i$-th nearest neighbor in $\mathcal{S}$
(under metric $d$).
Then LID can be approximated by
\[ \text{LID}_d(x) = - \left({1 \over k} \sum\limits_{i=1}^k \log {r_i(x) \over r_k(x)} \right)^{-1} \]
where $k$ is a defense hyperparameter the controls the number of
nearest neighbors to consider.
The authors use the distance function
\[d_j(x,y) = \left\lVert{}f^{1..j}(x) - f^{1..j}(y)\right\rVert{}_2\]
to measure the distance between the $j$th activation layers.
The authors compute a vector of LID values for each sample:
\[\overrightarrow{\text{LID}}(x) = \{\text{LID}_{d_j}(x)\}_{j=1}^n.\]
Finally, they compute the $\overrightarrow{\text{LID}}(x)$
over the training data and adversarial examples generated
on the training data,
and train a logistic regression classifier to detect
adversarial examples.
We are grateful to the authors for releasing their complete
source code.

\paragraph{Discussion.}
While LID is not a defense itself, the authors assess the ability of
LID to detect different types of attacks.

Through solving the formulation
\[ \text{min. } |x-x'|_2^2 + \alpha\left(\ell(x') + \text{LID-loss}(x')\right) \]
the authors attempt to determine if the LID metric is a good metric for
detecting adversarial examples. Here, $\text{LID-loss}(\cdot)$ is a function that can be minimized to
reduce the LID score.
However, the authors report that this modified attack still achieves $0\%$
success.
Because Carlini and Wagner's $\ell_2$ attack is unbounded,
any time the attack does not reach $100\%$ success indicates
that the attack became stuck in a local minima.
When this happens, it is often possible to slightly modify the loss
function and return to $100\%$ attack success \cite{carlini2017magnet}.

In this case, we observe the reason that performing this type of adaptive
attack fails is that gradient descent does not succeed in optimizing the LID
loss, even though the LID computation is differentiable.
Computing the LID term involves computing the $k$-nearest
neighbors when computing $r_i(x)$.
Minimizing the gradient of the
distance to the current $k$-nearest neighbors
is not representative of the true direction to travel in for
the optimal set of $k$-nearest neighbors.
As a consequence, we find that adversarial examples generated
with gradient methods when
penalizing for a high LID either (a) are not adversarial; or (b)
are detected as adversarial, despite penalizing for the LID loss.

\paragraph{Evaluation.}
We now evaluate what would happen if a defense would directly
apply LID to detect adversarial examples.
Instead of performing gradient descent over a term that is
difficult to differentiate through, we have found that
generating high confidence adversarial examples
\cite{carlini2017adversarial}
(completely
oblivious to to the detector)
is sufficient to fool this detector.
We obtain from the authors their detector trained
on both the Carlini and Wagner's $\ell_2$ attack and train our own
on the Fast Gradient Sign attack, both of which were found
to be effective at detecting adversarial examples generated
by other methods.
By generating high-confidence adversarial examples
minimizing $\ell_\infty$ distortion, we are able to reduce model
accuracy to $2\%$ success within $\epsilon=0.015$.
LID reports these adversarial examples are benign at a
$97\%$ rate (unmodified test data is flagged as benign with a
$98\%$ rate).

This evaluation demonstrates that the LID metric can be
circumvented, and future work should carefully evaluate if
building a detector relying on LID is robust to adversarial examples
explicitly targeting such a detector.
This work also raises questions whether a large LID is a fundamental
characteristic of all adversarial examples, or whether it is a by-product
of certain attacks.

\section{Defense-GAN}
\label{app:dgan}

\paragraph{Defense Details.}
The defender first trains a Generative Adversarial Network
with a generator $G(z)$ that maps
samples from a latent space (typically $z \sim \mathcal{N}(0,1)$)
to images that look like training data.
Defense-GAN
takes a trained classifier $f(\cdot)$,
and to classify an input $x$, instead of returning $f(x)$,
returns $f(\mathop{\text{arg min}}_z |G(z)-x|)$.
To perform this projection to the manifold, the authors take
many steps of gradient descent starting from different
random initializations.

Defense-GAN was not shown to be effective on \cifar{}.
We therefore
evaluate it on MNIST (where it was argued to be secure).

\paragraph{Discussion.}
In \citet{samangouei2018defensegan}, the authors construct a white-box attack by unrolling
the gradient descent used during classification.
Despite an unbounded
$\ell_2$ perturbation size, Carlini and Wagner's attack only reaches
$30\%$ misclassification rate on the most vulnerable model
and under $5\%$ on the strongest.
This leads us to believe
that unrolling gradient descent breaks gradients.

\paragraph{Evaluation.}
We find that adversarial examples \emph{do} exist on the data manifold as described
by the generator $G(\cdot)$.
However, Defense-GAN \emph{does not} completely project to the
projection of the generator, and therefore often does not identify these
adversarial examples actually on the manifold.

We therefore present two evaluations. In the first, we assume that
Defense-GAN were to able to perfectly project to the data manifold,
and give a construction for generating adversarial examples.
In the second, we take the actual implementation of Defense-GAN as it
is, and perform BPDA to generate adversarial examples with $50\%$ success
under reasonable $\ell_2$ bounds.

\bigskip
\noindent
\textbf{Evaluation A.}
Performing the manifold projection
is nontrivial as an inner optimization step when generating
adversarial examples.
To sidestep this difficulty,
we show that adversarial examples exist \emph{directly on}
the projection of the generator.
That is, we construct an adversarial example
$x' = G(z^*)$ so that
$|x-x'|$ is small and $c(x) \ne c(x')$.

To do this, we solve the re-parameterized formulation
$$\text{min. } \; \|G(z) -x\|_2^2 + c \cdot \ell(G(z)). $$
We initialize $z = \mathop{\text{arg min}}_z |G(z)-x|$
(also found via gradient descent).
We train a WGAN using the code the authors provide
\cite{gulrajani2017improved},
and a MNIST CNN to $99.3\%$ accuracy.

\begin{figure}
  \center
  \includegraphics[scale=.125]{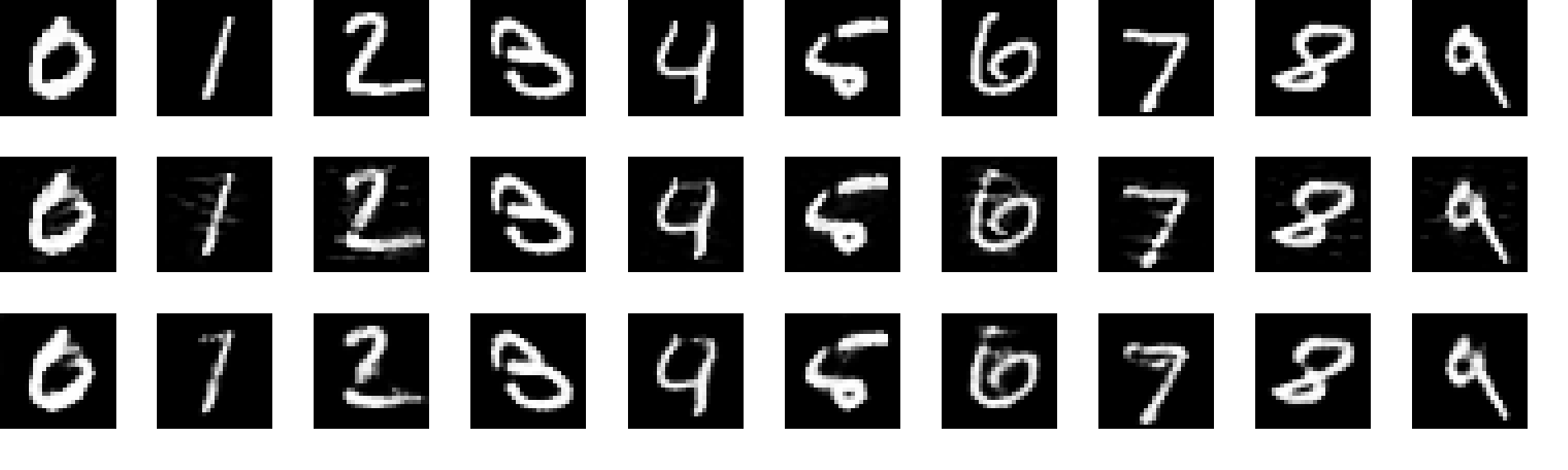}
  \caption{Images on the MNIST test set.
    Row 1: Clean images. Row 2: Adversarial examples on an unsecured classifier.
    Row 3: Adversarial examples on Defense-GAN.}
  \label{fig:defgan}
\end{figure}

We run for 50k iterations of gradient descent for generating
each adversarial example; this takes under one minute per instance.
The unsecured classifier requires a mean $\ell_2$ distortion of $0.0019$
(per-pixel normalized, $1.45$ un-normalized) to fool.
When we mount our attack, we require a mean distortion
of $0.0027$, an increase in distortion of $1.46\times$; see
Figure~\ref{fig:defgan} for examples of adversarial examples.
The reason our attacks succeed with $100\%$ success
without suffering from vanishing or exploding gradients
is that our gradient computation only needs to differentiate
through the generator $G(\cdot)$ once.

Concurrent to our work, \citet{ilyas2017robust} also develop a nearly
identical approach to Defense-GAN; they also find it is vulnerable to the
attack we
outline above, but increase the robustness further with adversarial training.
We do not evaluate this extended approach.

\bigskip
\noindent
\textbf{Evaluation B.}
The above attack \emph{does not} succeed on Defense-GAN.
While the adversarial examples \emph{are} directly on the projection of the Generator, the
projection process will actually move it \emph{off} the projection.

To mount an attack on the approximate projection process, we use the
BPDA attack regularized for $\ell_2$ distortion.
Our attack approach is identical to that of PixelDefend, except we
replace the manifold projection with a PixelCNN with the manifold
projection by gradient descent on the GAN.
Under these settings, we succeed at reducing model accuracy to $55\%$
with a maximum normalized distortion of $.0051$ for successful attacks.

\end{document}